# From Knowledge Organization to Knowledge Representation and Back

Fausto Giunchiglia[1], Mayukh Bagchi[1] and Subhashis Das[2]

1. DISI, University of Trento, Trento, Italy.
   Email: {fausto.giunchiglia, mayukh.bagchi}@unitn.it

2. CeIC, ADAPT, Dublin City University, Dublin, Ireland.
   Email: subhashishhh@drtc.isibang.ac.in

## Abstract
Knowledge Organization (KO) and Knowledge Representation (KR) have been the two mainstream methodologies of knowledge modelling in the Information Science community and the Artificial Intelligence community, respectively. The facet-analytical tradition of KO has developed an exhaustive set of guiding canons for ensuring quality in organising and managing knowledge but has remained *limited* in terms of *technology-driven activities* to expand its scope and services beyond the bibliographic universe of knowledge. KR, on the other hand, boasts of a robust ecosystem of *technologies* and *technology-driven service design* which can be tailored to model any entity or scale to any service in the entire universe of knowledge. This paper elucidates both the facet-analytical KO and KR methodologies in detail and provides a functional mapping between them. Out of the mapping, the paper proposes an integrated KR-enriched KO methodology with all the standard components of a KO methodology plus the advanced technologies provided by the KR approach. The practical benefits of the methodological integration has been exemplified through the flagship application of the Digital University at the University of Trento, Italy.

## Introduction
Knowledge Organization (KO) has been understood as comprising the functional triad of roles, artefacts and activities involved in the *"description, indexing and classification"*[1] of information resources (e.g., books) in memory institutions like libraries. In the context of the rich scientific literature condensed over almost a century now, KO research enquiry has chiefly focused on two aspects, namely, *KO Systems* and *KO Approaches*. KO systems[2] encompass a diverse variety of schemes for organising and representing knowledge of a subject, including, amongst others, classification schemes, cataloguing codes, and so on. KO approaches focus on ordered combinations of different KO systems towards achieving the most helpful sequence of organising knowledge, based on an underlying set of (implicit or explicit) guiding principles[3,4]. Knowledge Representation (KR), on the other hand, has been developed within the Artificial Intelligence (AI) community to investigate *"how knowledge can be represented symbolically"*[5] in intelligent systems. The mainstream KR research community has mainly concentrated on developing an enormous body of research in two aspects, *viz.*, the development of *KR Artefacts* and the design of *KR Methodologies*. KR artefacts encompass a diverse range of computational models for representing (domain) knowledge in AI systems including, for instance, ontologies[6], Knowledge Graphs (KGs)[7]. KR methodologies, on the other hand, focus on the step-by-step elucidation of roles and activities crucial to the development of KR artefacts for diverse application purposes, such as, data integration[8] and image annotation[9,10]. It should be noted that while KR has historically laid less emphasis on the development of guiding principles for modelling, it has excelled in terms of developing an advanced suite of *technologies* and *technology-driven service design* (e.g., the Semantic Web technology stack) which can be tailored to model any entity or scale to any service in the entire universe of knowledge.

Notice two dimensions which are apparent from the brief overviews on KO and KR given above. First, there is a similarity between KO and KR in terms of their high-level *functionality*, i.e., roles, artefacts, and activities, to methodologically design and develop models which can organise and represent knowledge for a domain. Second, KO and KR have complementary strengths and weaknesses, with KO being strong on guiding principles and weak in terms of technology-driven activities and services and vice versa for KR. Keeping in mind the preceding dimensions, this current work concentrates on two specific research questions. First, we enquire whether there is any high-level functional correspondence between a mainstream facet-analytical KO methodology and a mainstream KR methodology, respectively. Second, we ask whether an integrated methodology can be developed with all the strengths of KO *plus* the strengths of KR. Throughout this paper, by KO, we assume and concentrate on the mainstream facet-analytical approach originally postulated and practised by Ranganathan[3,4]. This is not by chance. There are three main reasons for our choice here. On one hand, Ranganathan's approach and proposed methodology, being highly structured, is naturally amenable to *algorithmisation*. On the other hand, Ranganathan has developed a huge body of high-quality knowledge management guiding principles (e.g., canons) that integrates well with work in KR, which, as noted above, tends to be rather weak in this dimension. Indeed, there already exist examples of how Ranganathan's heritage has been exploited towards the development of more well founded KR technologies and methodologies[11]. Last but not least, KO has remained *limited* in terms of *technology-driven activities*, thereby, minimising the otherwise extended space of entities and services it potentially could serve beyond the bibliographic universe of knowledge, if it fully exploited the technology and technology-based services advanced by KR.

In response to the above research questions, this paper proposes a solution strategy in two steps. First, it illustrates and elaborates on the roles, artefacts, and activities which compose the high-level view of a KO and KR methodology, respectively. Second, the paper presents a functional mapping between the components of the above two methodologies, and, by exploiting the mapping, proposes an integrated KR-enriched KO methodology that subsumes all the standard components of KO methodology with their scope and services enhanced many-fold via the incorporation of advanced technologies and services from KR methodology. To that end, the paper describes the characteristic features of the Digital University (DU)[12] infrastructure used at the University of Trento, Italy, as a flagship application based on an (adapted) implementation of the proposed integrated methodology. The discussion especially concentrates on and contrasts the limitations of the static scope and services offered by a mainstream Digital Library with the technology-enhanced purpose-specific services that can be customised for a specific DU implementation.

The remainder of the paper is organised as follows. The following section elaborates the research questions tackled by the paper via a concrete motivating example and insights from related work in KO and KR. The third section of the paper describes the components which compose the different phases of Ranganathan's facet-analytical KO methodology. The next section, in turn, describes the components within the different phases of a standard KR methodology. The fifth section of the paper performs a functional mapping between the components of the above methodologies and integrates them to propose an integrated KR-enriched KO methodology. The section following it describes the immediate benefits of a KR-enriched KO methodology by describing the notion of a DU and highlighting the services of the DU of the University of Trento, Italy. A final section gives some general conclusions.

**Research Questions**
Let us begin with a motivating example illustrated in Figure 1 depicting two different representations of the description of one and the same entity: in this case a book entitled *Bibliography on the tropical disease of children in India in the 1970s,* which was written by the author *E.F, Schumacher* and published in the year *1973*. On the top half of the left-hand side of Figure 1, we notice a "sample" card from a card catalogue describing the book. It displays imprint metadata for the book, namely, the title, author, publisher, place of publication, date of publication, and number of pages, in a way that follows the specifications of a catalogue code. It also contains two additional information: the accession number and the call number of the book. The rest of Figure 1 depicts (a fragment of) a sample Entity Graph (EG),[8] which is a KG visualisation of the description of the book modelled via nodes and edges. For example, the entity *book* (i.e., the red-coloured node) is related to the entities *E.F. Schumacher* and *Harper & Row* via the edges *'author'* and *'publisher'*, respectively. Worth noting are two additional features of the EG view of the book over and above the common metadata from the card catalogue card (depicted partially in the EG). First, several entities (e.g., *Harper & Row*) are classified in terms of the general *ontological* category to which they belong, or, in other words, their *entity types*[13] (e.g., *Harper & Row* is of the entity type *Organization*). This ontological hierarchy through which the EG is semantically structured is (partially) depicted in the graph labelled as Entity Type Graph (ETG)[8] in Figure 1. Second, one should also note that the EG also encodes a lot of extra metadata about the entities encoded as imprint information in the card catalogue card. For example, while *Harper & Row* was just mentioned as the publisher in the card catalogue record, the EG additionally informs the user that it is of the type *Organization*, headquartered in *Manhattan* and founded by *James Harper*. The panel on the right hand side of Figure 1 provides a partial view of some selected metadata about the EG itself, e.g., its IRI, timestamp, and so on.

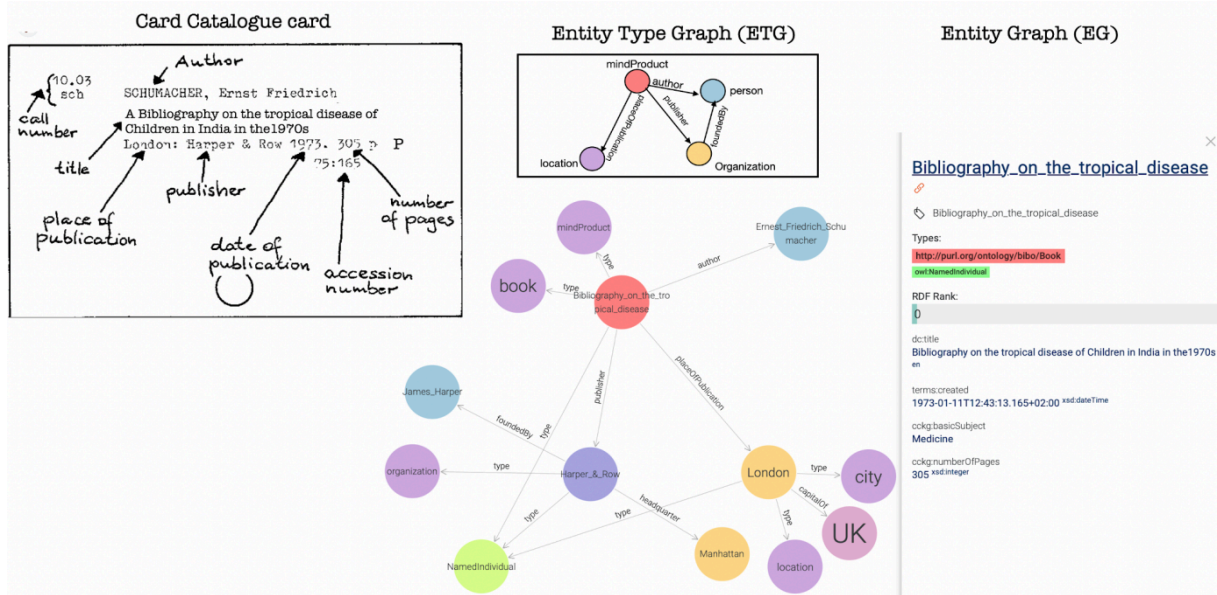

**Figure 1: A motivating example.**

The first research question is inspired by the nature of informational overlap (common metadata) and non-overlap (extra metadata) in the description of the book as encoded by the card catalogue record (a KO artefact) and the EG (a KR artefact) as exemplified above. It can be formulated as follows: *Is there a functional correspondence between the high-level views of KO and KR methodologies?* From the perspective of informational overlap, this research question is recursively dependent on several sub-questions. First, there are questions as to whether the methodologies of KO and KR are composed of constituent phases, and if so, whether there is a functional correlation between corresponding phases of KO and KR methodologies. Second, within each corresponding phase of KO and KR, one may ask whether there is a functional correlation amongst their constituent roles, artefacts, and activities. From the perspective of informational non-overlap, the research question should address the comparative scope and, thereafter, the purpose emphasised by both the KO and KR methodologies.

The second research question depends on the answer to the first research question. To that end, it can be formulated as follows: *Is a functionally integrated methodology with all the strengths of KO plus the technological advantages of KR feasible?* This question, too, is recursively dependent on some sub-questions. First, given a high-level functional correlation between the corresponding phases of the KO and KR methodologies, Can they be collapsed into one single function-preserving phase in an integrated methodology? Second, given the functional correlation between constituent components (roles, artefacts, and activities) of two corresponding phases, can the components be collapsed in a completely function-preserving manner? If not, how does this impact the functionally integrated methodology? Further, how should the scope and the purpose of the functionally integrated methodology be defined, These questions hold the key to determining the feasibility of a single artefact, functionally identical to the card catalogue record as well as the EG, representing all of the information in Figure 1 and extensible and exploitable as per requirements.

Finally, let us briefly concentrate on some related work relevant to the two research questions presented above. Here we wish to consider research done in KO focused on developing its methodological basis. Tennis (2008)[14] developed a metatheoretical framework of KO based on epistemology, theory, and methodology. His paper presented a high level view of KO methodologies but did not illustrate or explicate, step-by-step, any of the mainstream KO methodologies. Souza *et al.*'s (2012) highly cited work on the taxonomy of KOSs[15] proposed a hierarchy of intrinsic and extrinsic dimensions which should be integral to any KOS but didn't dwell on how they can be integrated into a well-founded KO methodology. López-Huertas (2008)[16] discussed several research dimensions crucial to next generation KO research and stressed the methodological development of KO beyond books to multidimensional knowledge and knowledge-based services. In sync with the above, Gnoli (2008)[17] proposed a list of long-term research questions in KO which include, amongst others, the need to investigate the expansion of the scope of KO and the need to pursue research on the technological enrichment of KO. It is also worth noting that researchers working on KR methodologies[18] have not attempted to chart out any form of mapping from KR to KO. Finally, Giunchiglia *et al.*'s (2021) research, which is based on the DERA methodology,[11] is complementary to the current work in that it proposes a detailed motivational basis and a set of modelling parameters following which KR methodologies can benefit from KO.

**Facet-Analytical Knowledge Organization Methodology**

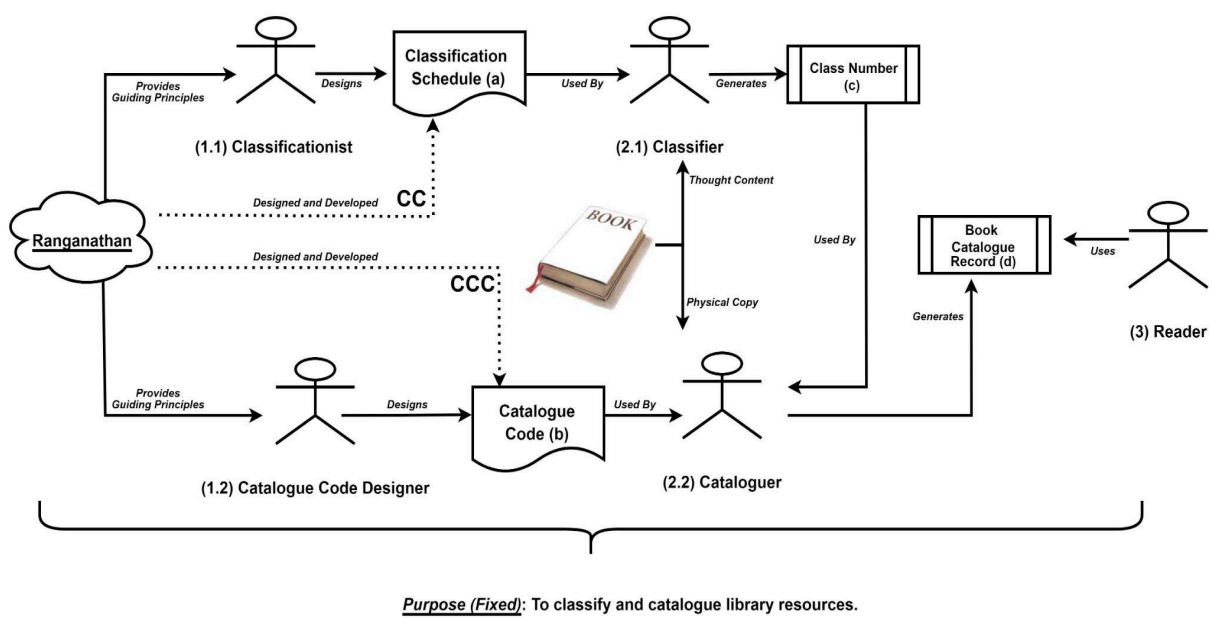

Figure 2: A high-level view of the facet-analytical KO methodology.

At the outset, let us elucidate the various phases, and in each phase, the various components, i.e., roles, activities, and artefacts, which compose to form a facet-analytical KO methodology whose chief purpose is to classify and catalogue library and information resources. Please consult Figure 2 for a high-level view of the methodology. The methodology consists of the following four distinct phases (the terminologies being detailed later):

1. The first phase entails the formulation of a classification schedule based on guiding principles[3] and the generation of a Class Number on the basis of that classification schedule.
2. The second phase involves the design of a catalogue code likewise based on guiding principles[4].
3. The third phase has to do with how the cataloguer integrates three different inputs, namely, the Class Number, the physical copy of the book, and the catalogue code, to generate the book catalogue record.
4. The final phase deals with how a reader visiting a library can use and exploit the book catalogue record.

The icons in Figure 2 include the roles (illustrated via the actor icon), activities (illustrated as edge labels) and artefacts (illustrated via any other icon). The numbers and lowercase alphabets which identify roles and artefacts, respectively, in Figure 2, are used later (in the fifth section of this paper) for establishing a functional correspondence with analogues in the KR methodology. We now describe the phases of the methodology sequentially.

The first phase begins with the *canons of classification* postulated by Ranganathan[3] to guide the formulation of high-quality classificatory structures for any subject. We enumerate below an overview of the three groups of canons of classification that are core to the methodology as illustrated in Figure 2:

1. Canons of the *Idea Plane*, which guide the modelling of concepts in a classification hierarchy based on their perceivable properties[19,20]. They include: canons about *characteristics*, canons about *succession of characteristics*, canons about *arrays,* and canons about *chains.*
2. Canons of the *Verbal Plane*, which guide the proper linguistic rendering of the concepts modelled following the canons of the Idea Plane, e.g., the canon of reticence.
3. Canons of the *Notational Plane*, which guide the assignment of a unique identifier for each linguistically labelled concept in the classification hierarchy, e.g., the canon of *synonym* and the canon of *homonym*.

The preceding sets of canons of classification are taken as input guidelines formulated by the *Classificationist* for designing faceted classification schedules that organise knowledge at different levels of abstraction — e.g., the Colon Classification (CC)[21,22] designed and developed by Ranganathan himself (indicated with dashed lines). Once the faceted classification schedule(s) are ready, the *Classifier* (see Figure 2) proceeds to generate the Class Number (Figure 2) for the subject of a book. In order to achieve this, the Classifier, first, takes as input the *a priori* designed classification schedules that provide him or her with the classification hierarchy (with each concept uniquely identified via an identifier) as well as the facet formula following which concepts can be combined to generate the Class Number. Secondly, (s)he also takes as input the thought content of the book to be classified. Finally, the Classifier follows the steps of Ranganathan's analytico-synthetic classification number generation procedure[23] to generate the Class Number which identifies its subject.

The second phase of the methodology (see Figure 2) deals with the design and development of a catalogue code. In sync with the first phase, it starts with the *canons of cataloguing* as input principles proposed by Ranaganathan[4] to guide high-quality description of any book (or, generally any library resource). We brief some of the canons. For example, the canon of *sought heading* prescribes that the attributes used to describe the access to a book in a catalogue should be strictly based in accordance with how a user is likely to approach the catalogue. Further, the canon of *consistence* prescribes that, for a type of library resource, the (metadata) attribute schema which composes its catalogue record should be consistent, unless otherwise stipulated by the canon of *context*. The canon of *local variation* prescribes a degree of flexibility in the creation of the headings of a catalogue record to provide access based on the local users' needs. These canons of cataloguing are taken as input guidelines by the *Catalogue Code Designer* to design a *Catalogue Code* — a documented standard of reusable specifications on how and what metadata properties should be accounted for to describe a specific bibliographic resource type. In fact, Ranganathan himself designed and developed a catalogue code which he named the Classified Catalogue Code (or, CCC; shown in Figure 2 via dashed lines).

The third phase has to do with how the *Cataloguer* (see Figure 2) uses the outputs of the previous two phases and the physical copy of the resource (e.g., a book) being catalogued to generate a Book Catalogue Record(s). In this phase, the Cataloguer receives as input the Class Number (output of the first phase) which models the subject of the book following the appropriate classification schedule and its facet formula. Given the Class Number, chain indexing[4] is executed to generate subject headings for the catalogue record. Further, the Cataloguer also receives two other inputs: the physical copy of the book which encodes its imprint information, and, the catalogue code (e.g., CCC) which specifies the exact set of allowed fields and syntax of how such imprint information should be encoded in the catalogue record for a particular type of resource. At this stage, the Cataloguer also uses the Class Number, (part of) the imprint information and fields from the accession register to create the unique call number. Thereafter, the Cataloguer integrates the subject headings together with the requisite imprint attributes and the call number to generate the Book Catalogue Record(s).

The fourth and final phase focuses on how library readers can use the book catalogue record. First, the reader can use it to (manually) search for and identify a book. He or she can also determine whether (s)he is looking for the correct topic by consulting the subject headings in the record. Second, in the digital information environment, the reader can use the catalogue record in an Online Public Access Catalog (OPAC), possibly enhanced with library discovery services[24], thereby, going beyond the traditional means of library search to include *limited* web-scale exploratory search, but still focussing on entities in the bibliographic universe of knowledge.

## Knowledge Representation Methodology

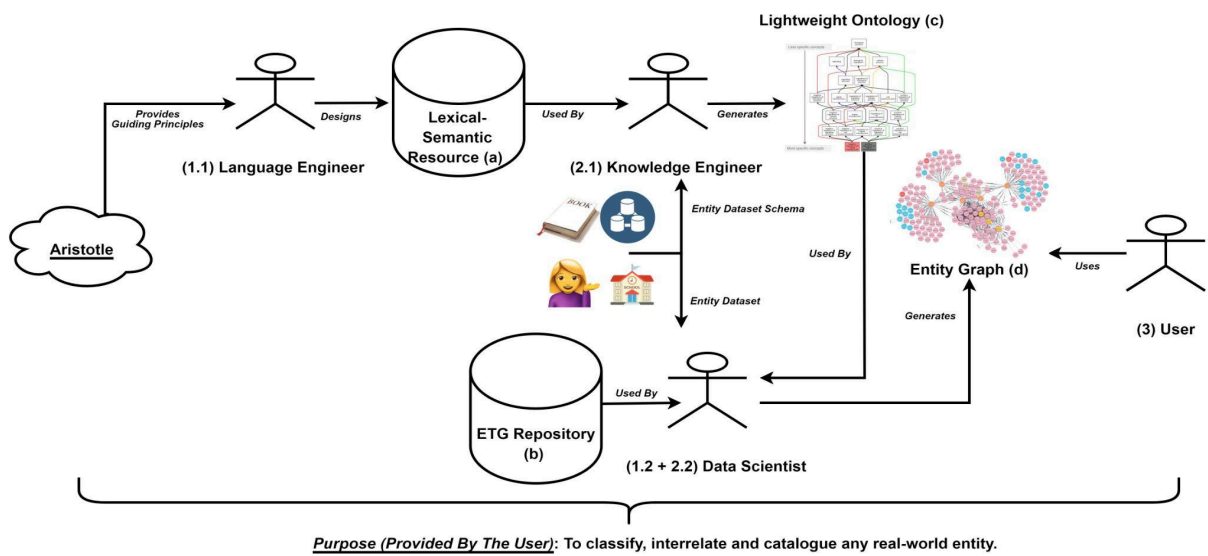

Figure 3: A high-level view of the KR methodology.

We shall now consider the various phases which compose to form a standard KR methodology, the goal of which is to classify, *interrelate* and catalogue any entity existing in the universe of knowledge in accordance with a purpose provided by a user. Figure 3 gives a high-level view of the KR methodology. This methodology comprises the following four distinct phases (the terminologies of which will be detailed later):

1. The first phase commences with guiding principles and concludes with the generation of the Lightweight Ontology[25].
2. The second phase focuses on the development of the Entity Type Graph (ETG) repository[26].
3. The third phase has to do with how a data scientist integrates three different inputs, namely, the Lightweight Ontology, Entity Dataset(s), and ETGs from the ETG repository to generate the Entity Graph (EG)[8].
4. The fourth phase concerns the different ways in which a user can use the EG.

The icons of Figure 3 are, for the most part, identical to those used in Figure 2.. We describe each phase of the methodology sequentially as follows.

The first phase commences with guiding principles articulated as the *Genus-Differentia*[27] paradigm for defining entities as established by by Aristotle[28]. According to the paradigm, any real-world entity denoted linguistically by a general noun can be defined by means of two categorical principles, the *Genus* and *Differentia*. A Genus defines an *a priori* set of properties shared across distinct entities — e.g., the property of being a stringed musical instrument: Differentia, on the other hand, defines a novel set of properties used to differentiate entities having the same Genus — e.g., the properties of musical instruments having three strings or six strings. These principles are taken as input by the *Language Engineer* (see Figure 3) to create machine-processable language datasets — e.g., WordNet-like[29,30] lexical semantic hierarchies encoding word meanings in different natural languages and representing different domains.

Such language datasets are hosted and managed via a *Lexical-Semantic Resource* (see Figure 3) which is usually designed as a collection of lexical-semantic hierarchies hosted by a repository and (publicly) accessible by a catalogue. Given the design of the lexical-semantic resource, the final activity of the first phase is implemented by the *Knowledge Engineer* (see Figure 3) who has to generate a *Lightweight Ontology* (Figure 3), an intermediate formal hierarchy *"consisting of backbone taxonomies"*[25] that is employed for representing knowledge for a particular application of the methodology. The Knowledge Engineer first takes as input the appropriate lexical-semantic hierarchy in a specific language which will guide the syntax and modelling of the taxonomic backbone of the lightweight ontology (s)he will generate. Further, (s)he also consults the schema of the datasets encoding the real-world entities which should be modelled in the lightweight ontology. This latter step facilitates in deciding the exact way in which the subsumption hierarchy of the lightweight ontology should be organised.

The second phase of the methodology is concerned with the design and development of a repository (aided by a catalogue) of reusable Entity Type Graphs (ETGs)[8,13]. ETGs are defined as formal ontological representations modelling entity types[13] and codifying the semantics inherent in (dataset) entities. An ETG, therefore, models *object properties* (i.e., properties that interrelate entities) and *data properties* (i.e., properties that describe entities) as core to any KR model. Notice, however, that the methodology (see Figure 3) *does not mandate any activity to guide the design of high-quality ETGs* composing the ETG repository. Once an ETG repository has been established, the third phase has to do with how a *Data Scientist* (see Figure 3) generates the final *Entity Graph* (EG)[8]. An EG is a Knowledge Graph (KG)[7] that is (i) taxonomically modelled via a lightweight ontology and (ii) supplied with object properties and data properties from a relevant ETG. In order to create an EG, the Data

Scientist receives as input the lightweight ontology (output of the first phase) which is then grounded in a relevant classificatorily-compliant ETG chosen from the ETG repository (output of the second phase). The resultant KR artefact, which constitutes a combination of the lightweight ontology and ETG, is now supplied with crucial object properties and data properties. The grounding also, non-trivially, ensures the completeness and reconfiguration[31] of the KR model in terms of unifying application specific concepts (modelled bottom-up) into general (domain) concepts (modelled top-down). Once the combined KR artefact has been created, the Data Scientist takes as input the entity datasets and semi-automatically maps[8] them, one at a time, to the KR artefact to generate the final EG.

Finally, the fourth and the last phase is concerned with how different users can exploit the EG. Different types of technology-driven services, from basic to intermediate to advanced, can be developed to explore the EG for this purpose. For instance, a set of basic services can include the option to download an EG (in full or a fragment thereof) in different formats which are directly exploitable in different AI tasks, such as, *OWL RDF/XML* for ontology engineering tasks, *FCA* format[26] for knowledge embedding tasks, and so on. An intermediate level of services can include, for instance, the option to query an EG via, e.g., the SPARQL query language and protocol, and options to produce different visualizations[26] of the EG. Finally, depending on the purpose enunciated by the user, any range of services can be customised to exploit the methodology and its various outputs. For instance, in a university, a smartphone application can be developed exploiting the methodology and the resultant university EG to deliver different information services to students via an integrated interface, such as timetables, canteen bookings, library borrowings, and so on. Notice that, in contrast to KO, KR provides an *exhaustive* set of technology-driven *services* and so on can be customised as per the requirements of users.

**The KR-enriched KO Methodology**

| KO Methodology | KR Methodology | KR-enriched KO Methodology |
|---|---|---|
| Classificationist (1.1) | Language Engineer (1.1) | Language Engineer (1.1) |
| Classifier (2.1) | Knowledge Engineer (2.1) | Knowledge Engineer (2.1) |
| Catalogue Code Designer (1.2)<br><br>Cataloguer (2.2) | Data Scientist (1.2 + 2.2) | Ontology Engineer (1.2 + 2.3)<br><br>Data Scientist (2.2) |
| Reader (3) | User (3) | User (3) |

**Table 1: Functional Mapping between roles of KO, KR and KR-enriched KO methodology.**

We are now in a position to establish a functional mapping of the roles and the artefacts of the two methodologies that we have discussed in this paper --the KO and KR methodologies (see the *first two columns* of Table 1). Three observations should be noted here. First, we ascertain, via the functional mapping, whether or not two roles or artefacts perform the same or synonymous broad *function*, irrespective of differences in their syntax or form. Second, in Table 1, we mention only the roles as they constitute the major functional difference between the two methodologies. The mapping of the artefacts is, on the other hand, presented only in the textual description as they are functionally synonymous. Thirdly, we perform the functional mapping in sync with the four informal phases characteristic of the two aforementioned methodologies. In the following explanation, we refer to Table 1, Figure 2, and Figure 3 as required.

In the first phase, there is a clear functional mapping between the roles of Classificationist (1.1) and that of Language Engineer (1.1). This is due to the fact that the function of both these roles is to guide the generation of (lexical) classification hierarchies on the basis of *a priori* input principles. Further, the artefacts produced by the above two roles - *viz.*, the Classification Schedule (a) (see Figure 2) and the Lexical-Semantic Resource (a) (see Figure (3) are also functionally similar, given that both are organised as a set of classification hierarchies focused on different domains. Furthermore, the roles of Classifier (2.1) and Knowledge Engineer (2.1) are also functionally similar because their core function is to use prescribed (lexical) classification hierarchies and classify the subjects of the resources they are tasked with describing. To that end, the output artefacts they produce — i.e., Class Number (c) (see Figure 2) and Lightweight Ontology (c) (see Figure 3) —play the same functional role of encoding the categorization of either a book's subject matter or an entity dataset schema.

In the second phase, the Catalogue Code (b) (see Figure 2) and the ETG Repository (b) (see Figure 3) are functionally mapped because both of these artefacts provide their input artefacts with reusable specifications regarding how to describe conceptual entities in a classification hierarchy via a set of properties. The first major discontinuity in the functional correspondence between the KO and KR methodologies occurs with respect to the role of the Data Scientist (1.2 + 2.2). The Data Scientist (KR) subsumes two KO roles, i.e., the Catalogue Code Designer (1.2) who specifies the descriptive (data) properties for a specific entity type and the Cataloguer (2.2) who generates the book catalogue record(s). On the other hand, the Book Catalog Record (d) (see Figure 2) and the Entity Graph (d) (see Figure 3) perform comparable functions in that they provide the end user with a unified view of top-down and bottom-up knowledge instantiated with data. Finally, the Reader (3) and User (3) are also fully analogous to one another as both function as end-users of the output of their respective methodologies. Important to note is the fact that the activities across both methodologies are also fully functionally similar (see the labelled edges on Figure 2 and Figure 3, respectively) with the sole exception being the lack of any agent associated with the design of the ETG repository in Figure 3.

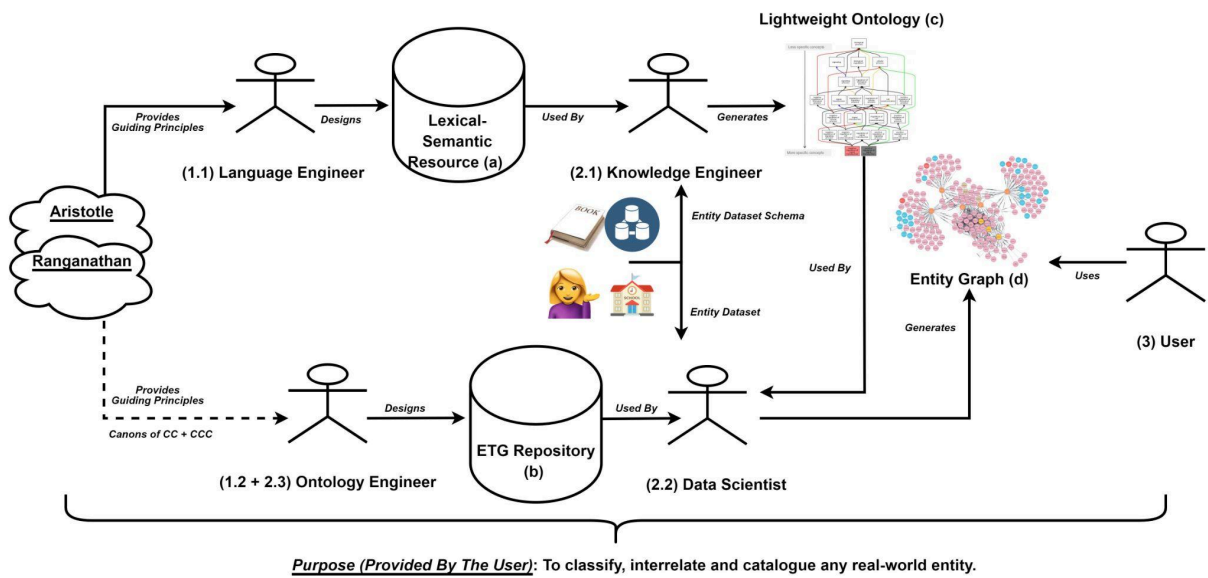

**Figure 4: The KR-enriched KO methodology.**

Having reviewed the mapping of agents across the KO and KR methodologies, we now turn to consider a new KR-enriched KO methodology that fully harmonises the functionally mapped features of both methodologies (see Figure 4). This new harmonised methodology is comprised of the following four distinct phases:

1. The first phase begins with guiding principles and concludes with the generation of the Lightweight Ontology.
2. The second phase involves the development of canons-based Entity Type Graphs (ETGs) constituting the ETG repository.
3. The third phase of the methodology is concerned with how the data scientist generates the Entity Graph (EG), and,
4. The final phase has to do with the different ways in which a user can use the EG.

Notice that this new KO methodology, technologically enriched and extended via KR, can be used to classify, describe and interrelate any bibliographic entity to any other bibliographic or non-bibliographic real-world entity in accordance with the specific purpose of a user . The icons of Figure 4 are identical to those used in Figures 2 and 3. We now describe the methodology phase-by-phase.

The first phase of the integrated methodology (see Figure 4) is the same as that of the original KR methodology (see Figure 3). The *Language Engineer* designs a *Lexical-Semantic Resource* composed of machine-processable lexical-semantic hierarchies in different languages and domains designed by following the *Genus-Differentia* paradigm. These lexical-semantic hierarchies are then used by the *Knowledge Engineer* together with entity dataset schemas to generate a *Lightweight Ontology*. The second phase of the methodology, i.e., the development of the ETG repository is markedly different from that in Figure 3. It begins with (the adaptation of) Ranganathan's canons of both classification and cataloguing as guides for the development of high-quality ETGs that make up the ETG repository. The use of the canons here is *fundamental* for the generation of taxonomically well-founded ETG hierarchies that are also encoded with canon-compliant data properties describing the (conceptual) entities contained within them. To that end, the methodology introduces the new role of an *Ontology Engineer*, whose task is to ensure the development of the ETGs, and thereby, the ETG repository, in full conformance with the quality guidelines prescribed by the Ranganathan's canons.

The rest of the methodology (see Figure 4) is also identical to the original KR methodology (see Figure 3). In the third phase, the Data Scientist integrates the lightweight ontology, the entity datasets, and the relevant canons-compliant ETG, in an iterative fashion, to generate an EG. The final phase has to do with the ways in which the user can exploit the output EG. One should note, however, the two powerful advantages that the integrated KR-enriched KO methodology has over the traditional facet-analytical KO methodology. First, the graph-theoretic technological basis of KR as instantiated in ontologies and ontology-based KGs expands the very scope of KO, namely, from being concerned with a *limited* set of bibliographic entities to a *countably infinite* set of any real-world entity. Second, the integrated methodology also benefits from the possibility of a diverse set of basic, intermediate, and advanced *purpose-specific service* designs which are a major advance on the otherwise limited set of (mostly manual) services which are constituent of the traditional KO methodology.

It is also worth considering how the new integrated methodology functionally maps to the KO and KR methodologies, respectively (see all columns of Table 1). The roles of the Language Engineer (1.1), Knowledge Engineer (2.1), and User (3) in the integrated methodology are functionally analogous to that of their equivalent counterparts in the original KR and KO methodologies (which we have already mapped to one another). The major addition is the role of the Ontology Engineer (1.2 + 2.3), who builds the classification canons-compliant ETG hierarchy supplied with cataloguing canons-compliant property attributions. The Data Scientist (2.2) in the integrated methodology is partially mapped to the role of Cataloguer in the KO methodology and to a part of the role of Data Scientist in the KR methodology, both of which are focused on the integration of three input artefacts to produce the final output artefact. As regards artefacts, the integrated methodology is functionally similar to the KO as well as the KR methodology. One may additionally observe that the integrated methodology also adds an extra activity path to ensure the canons-based quality compliance of the ETGs that undergirds the preparation of high-quality KR artefacts and contributes to the overall quality enhancement of the integrated KR-enriched KO methodology.

**From Digital Libraries to Digital Universities**

Let us now briefly consider Digital Libraries (DLs),[32] which are the products of a digital recast of the traditional process of knowledge organisation of bibliographic resources as depicted at a high-level in the (facet-analytical) KO methodology of Figure 2. The characteristics of DLs are as follows. First, depending on the scope of the subjects of the digital library resources, a content classification hierarchy is modelled, often featuring such categories as Communities (e.g., Information Science) which are hierarchically divided into collections of resources (e.g., articles, theses, presentations). Second, specific sets of metadata properties are stipulated for cataloguing each specific collection of resources. Thereafter, the cataloguer takes in three inputs, namely, the content classification hierarchy, the metadata schema, and the (digital) bibliographic entity, and suitably integrates them to generate a machine-readable catalogue record which can be consulted by potential users through an online (public) access catalogue (OPAC).

**Figure 5: A snapshot of the University of Trento Digital University portal.**

In contrast, Digital Universities (DUs)[33] are functionally based on the digitally-native, technology-driven process of KR-enriched knowledge organisation of resources as depicted at a high-level in the KR-enriched KO methodology of Figure 4. Within the framework of a DU, a preliminary lexical-semantic hierarchy is designed, which serves as the basis for modelling a lightweight ontology of the types of resources to be hosted in the DU. Then, the Data Scientist chooses a reference reusable ETG — that is to say, the DU ETG[12] — from the ETG repository. Thereafter, the Data Scientist takes as input the lightweight ontology, the (DU) ETG, and the concrete datasets of the (entity) resources to model the Digital University EG which is exploited by a variety of users in a variety of ways for a variety of purposes. Three observations should be noted here which make a DU distinct from a mainstream DL. First, a DU makes explicit the entity types of the entities it exposes in an underlying formal ontology which is not a necessary design component of a mainstream DL. To that end, the scope of DUs encompasses any real-world entity from bibliographic or archival resources to people, organisations, project support entities, legal entities, etc.: Indeed, it is completely customizable and individualised to the needs of its user community, for example, a university. This step is key to facilitate semantic interoperability of (meta)data coming from different DUs. Second, the population of a DU ontology results in an explicit DU KG integrating data of diverse entities interrelated amongst each other which, again, is not a necessary design component of a mainstream DL. This step is key to facilitate stratified data integration within and across DUs. Third, leveraging the underlying ontology and KG, the services of a DU not only cater to different university communities — such as, e.g., researchers (e.g., browsing information about other researchers and their linked publications and projects) or students (a DU-based mobile app for university study management) — but also to specialised extra-academic communities: for example, annual snapshots of a DU EG can be exposed via open data catalogues as open data, thereby enhancing the academic and administrative transparency of a university to all interested parties.

Let us consider a concrete example of a DU, namely, the Digital University[1] of the University of Trento (UniTn), Italy, which is publicly accessible and explorable (see Figure 5). From the very outset, the integrated KR-enriched KO methodology proposed in this paper has been adapted and implemented as a hub-and-spoke system architecture forming the methodological core of the UniTn DU. The UniTn DU has developed an expandable DU ETG that encodes concepts such as person, organisation, computer file, publication, creative work, course, project, event, and so on. This DU ETG is multilingual as it is based on the Universal Knowledge Core (UKC),[30,34] which is a powerful multilingual lexical-semantic resource hosted by the UniTn. Moreover, an automatic extract, transform, and load (ETL) procedure is executed at regular intervals to extract requisite information about relevant entities from the heterogeneous set of data sources, which, are then mapped to the UniTn DU ETG to generate a temporally-boxed snapshot of the UniTn DU EG. One should note that the UniTn DU EG integrates data and knowledge from databases as heterogeneous as the university people database, experts and competences database, statutory bodies databases, organisational units database, etc.

The general procedure followed to customise a service in a DU[12] and the services consequently made available by UniTn DU are as follows. At the outset, the focus is on consolidating the service requirements in terms of the target group(s), functionalities, and the data sources involved. Given the requirements, the next point of emphasis is to determine the target language(s) of the service (for example, this might require extending the UKC) and the consequent localization of the ETG with respect to the conceptual entities they are core to the requirements. After the consolidation of the language and ETG, the focus shifts to selecting the data sources and setting up the ETL procedure with respect to the sources to generate an updated DU EG. Finally, the service is concretely implemented and deployed by accessing the EG via customised APIs. It should not escape attention that the above step-by-step service design approach is scalable, cost-effective, and generalizable to any use case[12]. For example, see Figure 5 for a snapshot of the DU *institutional portal* of the UniTn. In it, the DU profile of UniTn faculty member Fausto Giunchiglia is shown (with the data, in this case, coming from the people database). It gives basic information about his position at the university and his research competences. It also lists his teaching activities, publications (shown in Figure 5), theses supervised by him, research projects, and office hours. This profile, of course, is only a small fragment (or, in information science parlance, a catalogue record) of the overall DU EG. Currently, the UniTn DU provides a total of four key services[12], namely:
1. *UniTn Institutional Portal*, as exemplified and described above via Figure 5;
2. *UniTn Institutional Dashboard*, a data analytics service providing multidimensional assessment of the research conducted at the university, thereby facilitating improvement in research governance, funding, and management;

---
[1] https://webapps.unitn.it/du/en

3. *UniTn Open Data Publication*, a broader data interoperability service that publishes periodic snapshots of the UniTn DU EG on regional, national, and European open data portals.
4. *UniTn Mobile App*, a multidimensional university study-centred personalised activity management service for students.

**Conclusion**

The paper has detailed two novel contributions to the field of KO. First, it has elucidated a high-level functional mapping between the facet-analytical KO methodology and the KR methodology by illustrating each methodology in detail. Second, based on the functional mapping, it has proposed an integrated KR-enriched KO methodology with all the components of KO plus the technological advantages of KR. It has also contextualised the practical advantages of the KR-enriched KO methodology in context by presenting brief highlights of the flagship Digital University infrastructure of the University of Trento in Italy.

**Acknowledgement**

With this paper, we would like to pay our obeisance to S.R. Ranganathan who left a lasting impact on the theory and practice of classification and cataloguing as we know today. We acknowledge the early collaborative work by Dr. Biswanath Dutta, Dr. A.R.D. Prasad and Dr. Devika Madalli in developing the foundations and methodologies which led to the fruition of the current work. Finally, we thank our colleagues Simone Bocca and Alessio Zamboni for their key contribution in implementing the technologies described in the work.

The work described in this paper follows on and completes the work described in a twin paper by Giunchiglia and Bagchi (2024)[35]. The focus of this latter paper goes somehow in the opposite direction, being on how the guiding principles of classification and cataloguing can be leveraged and integrated to model high quality KR artefacts. The focus application is the high quality annotation of images used for training computer vision (CV) systems. We refer to this companion methodology as the KO-enriched KR methodology. Figures 2, 3, 4 and Table 1 were first introduced in Giunchiglia and Bagchi (2024)[35] and are used here with minor modifications.